\documentclass[runningheads]{llncs}
\usepackage{graphicx}
\usepackage[dvipsnames]{xcolor}
\usepackage{tikz}
\usepackage{comment}
\usepackage{amsmath,amssymb} % define this before the line numbering.
\usepackage{color}
\usepackage{graphicx}
\usepackage{amsmath}
\usepackage{amssymb}
\usepackage{booktabs}
\usepackage{xspace}
\usepackage{bm}
\usepackage{physics}
\usepackage{multirow}
\usepackage{enumitem}
\usepackage{makecell}
\usepackage[pagebackref,breaklinks,colorlinks]{hyperref}

\usepackage[accsupp]{axessibility}  % Improves PDF readability for those with disabilities.
\usepackage[capitalize]{cleveref}
\crefname{section}{Sec.}{Secs.}
\Crefname{section}{Section}{Sections}
\Crefname{table}{Table}{Tables}
\crefname{table}{Tab.}{Tabs.}

\def\eg{\emph{e.g.}} 
\def\ie{\emph{i.e.}}

\def\etal{\emph{et al.}}

\begin{document}
% \linenumbers
\pagestyle{headings}
\mainmatter
\def\ECCVSubNumber{5118}  % Insert your submission number here

\title{Compound Prototype Matching for Few-shot Action Recognition} % Replace with your title

\begin{comment}
\titlerunning{ECCV-22 submission ID \ECCVSubNumber} 
\authorrunning{ECCV-22 submission ID \ECCVSubNumber} 
\author{Anonymous ECCV submission}
\institute{Paper ID \ECCVSubNumber}
\end{comment}
%******************

% \begin{comment}
\titlerunning{Compound Prototype Matching}
\author{Yifei Huang \and
Lijin Yang \and
Yoichi Sato}
\authorrunning{Huang et al.}
\institute{Institute of Industrial Science, the University of Tokyo, Tokyo, Japan
\email{\{hyf,yang-lj,ysato\}@iis.u-tokyo.ac.jp}}
% \end{comment}
%******************
\maketitle

\begin{abstract}
Few-shot action recognition aims to recognize novel action classes using only a small number of labeled training samples. 
In this work, we propose a novel approach that first summarizes each video into compound prototypes consisting of a group of global prototypes and a group of focused prototypes, and then compares video similarity based on the prototypes. Each global prototype is encouraged to summarize a specific aspect from the entire video, \eg, the start/evolution of the action. Since no clear annotation is provided for the global prototypes, we use a group of focused prototypes to focus on certain timestamps in the video. 
We compare video similarity by matching the compound prototypes between the support and query videos. The global prototypes are directly matched to compare videos from the same perspective, \eg, to compare whether two actions start similarly. For the focused prototypes, since actions have various temporal variations in the videos, we apply bipartite matching to allow the comparison of actions with different temporal positions and shifts.
Experiments demonstrate that our proposed method achieves state-of-the-art results on multiple benchmarks. 
\end{abstract}

\section{Introduction}
\label{sec:intro}
Difficulty in collecting large-scale data and labels promotes the research on few-shot learning. 
Built upon the success of few-shot learning for image understanding tasks~\cite{li2019finding,chowdhury2021few,xu2021learning,kang2019few,fan2020few,wang2019panet,liu2020crnet,liu2020part,wang2020few}, many begin to focus on the few-shot action recognition task~\cite{kliper2011one}. 
Once realized, such techniques could greatly alleviate the cost of video labeling~\cite{grauman2021ego4d} and promote real-world applications where the labels in certain scenarios are hard to acquire~\cite{huang2018predicting}. 

Most few-shot action recognition works determine the category of a query video using its similarity to the few labeled support videos. Many works~\cite{zhu2020label,zhu2021few,fu2020depth} follow  ProtoNet~\cite{snell2017prototypical} to first learn a prototype for each video and compute the video similarity based on the similarity of the prototypes. 
To better consider the temporal dependencies in videos (\eg, temporal ordering and relation), recent works construct sub-sequence prototypes using different parts of the videos and calculate video similarity by matching the support-query prototypes~\cite{cao2020few,perrett2021temporal}. 

While temporal dependencies are considered, limitations still exist in previous methods. Firstly, without considering spatial information, previous methods cannot fully exploit the spatiotemporal relation in the videos for distinguishing actions like ``put A on B" and ``put A besides B" because they differ only in the relative position of the objects. Secondly, the sub-sequence prototypes come from fixed temporal locations, so that they cannot well handle the actions that happen at different speeds. Thirdly, it is computationally costly to exhaustively compute the similarity between all pairs of sub-sequence prototypes~\cite{perrett2021temporal}.

To address the limitations and achieve more robust few-shot action recognition, we explore how to: (1) better generate prototypes that can robustly encode spatiotemporal relation in the videos, (2) enable the prototypes to flexibly encode the actions done with different lengths and speeds, and (3) match the prototypes between two videos without exhaustively comparing all prototype pairs. 

One straightforward way to address the first point is to use object features extracted from object bounding boxes~\cite{he2017mask}. However, we observed in our preliminary experiments only limited performance gain when previous methods~\cite{perrett2021temporal,zhang2021learning} directly use them as additional inputs. In this work, we propose a multi-relation encoder to effectively encode the object features, by considering the spatiotemporal relation among objects across frames, the temporal relation between different frame-wise features, as well as the relation between object features and frame-wise features.

For the second and third points, we propose to generate \textit{global prototypes} that consider all frames in the input video. This is done by taking advantage of the self-attention mechanism of Transformers~\cite{vaswani2017attention}. Since it is difficult to represent a wide variety of actions by using a single prototype, we instead use a group of prototypes to represent each action. 
We match the support-query similarity by fixed 1-to-1 matching, \ie, the $i$-th prototype of the query video is always matched with the $i$-th prototype of the support video. Thus, during training, each prototype will try to capture a certain aspect of the video. To avoid all prototypes to be the same, we apply a diversity loss when learning the prototypes, so that they are encouraged to capture different aspects of the action (\eg, one prototype captures the start of the action, and another prototype captures the action evolution). 

However, learning prototypes to represent different aspects of the action (\eg, the start/end of the action) is a difficult task even with annotation~\cite{chao2018rethinking,luo2020weakly,ma2021weakly,huang2020improving}. Thus, it is not sufficient to compute video similarity only by the global prototypes. To make the similarity measurement more reliable, we generate another group of \textit{focused prototypes}, where each prototype is encouraged to focus on certain timestamps of the video. 
Since actions may happen in different parts of the videos at different speeds~\cite{li2021ta2n}, it is not correct to compare the same timestamp between two videos.
We therefore use bipartite matching to match the focused prototypes between the support and query videos, so that comparison of actions with different lengths and speeds can be done. 

Our method uses the compound of two groups of prototypes, which we call \textit{compound prototypes}, for calculating video similarity. On multiple benchmark datasets, our method outperforms previous methods significantly when only one example is available, demonstrating the superiority of using our compound prototypes in similarity measurement.

To summarize, our key contributions include:
\begin{itemize}
    \item A novel method for few-shot action recognition based on generating and matching compound prototypes.
    \item Our method achieves state-of-the-art performance on multiple benchmark datasets~\cite{carreira2017quo,goyal2017something,kuehne2011hmdb}, outperforming previous methods by a large margin.
    \item A detailed ablation study showing the usefulness of leveraging object information for few-shot action recognition and demonstrating how the two groups of prototypes encode the video from complementary perspectives.
\end{itemize}

%-------------------------------------------------------------------------
\section{Related Works}
\textbf{Few-shot image classification} methods can be broadly divided into three categories. The transfer-learning based methods~\cite{dhillon2019baseline,qiao2018few,wang2020frustratingly} use pre-training and fine-tuning to increase the performance of deep backbone networks on few-shot learning.
The second line of works focuses on rapidly learning an optimized classifier using limited training data~\cite{andrychowicz2016learning,gui2018few,ravi2016optimization,antoniou2018train,wei2019piecewise,finn2017model,zhang2020deepemd}. The third direction is based on metric learning, whose goal is to learn more generalizable feature embeddings under a certain distance metric~\cite{koch2015siamese,snell2017prototypical,vinyals2016matching,kang2021relational}. The key to metric-learning based methods is to generate robust representations of data under a certain metric, so that it may generalize to novel categories with few labeled samples.
Our work falls into this school of research and focuses on the more challenging video setting. 

\textbf{Few-shot action recognition} methods~\cite{thatipelli2021spatio,li2021ta2n,patravali2021unsupervised,hong2021video,bishay2019tarn,zhu2021few,mishra2018generative,zhu2021closer,xu2018dense} mainly fall into the metric-learning framework. Many works follow the scheme of ProtoNet~\cite{snell2017prototypical} to compute video similarity based on generated prototypes. For learning better prototypes, ProtoGAN~\cite{kumar2019protogan} synthesizes additional feature, CMN~\cite{zhu2018compound,zhu2020label} uses a multi-layer memory network, while ARN~\cite{zhang2020few} uses jigsaws for self-supervision and enhances the video-level representation via spatial and temporal attention. There are also methods that perform pretraining with semantic labels~\cite{wang2021semantic,xian2020generalized,xian2021generalized} or use additional information such as depth~\cite{fu2020depth} to augment video-level prototypes. Our method uses another form of additional information: the object bounding box from pretrained object detectors~\cite{he2017mask}.

% Since videos add an additional temporal dimension compared with images, directly comparing video-level prototypes may match the misaligned actions, leading to a sub-optimal similarity measurement.
The temporal variance of actions form a major challenge in few-shot action recognition task~\cite{li2021ta2n}.
To better model temporal dependencies, recent researches put more focus on generating and matching sub-video level prototypes.
OTAM~\cite{cao2020few} uses a generalized dynamic time warping technique~\cite{chang2019d3tw} to monotonously match the prototypes between query and support videos.
ITA-Net~\cite{zhang2021learning} first implicitly aggregates information for each frame using other frames, and conducts a 1-to-1 matching of all prototypes.
However, these two methods use frame-wise prototypes, thus cannot well capture the higher-level temporal relation in multiple frames.
Recently, TRX~\cite{perrett2021temporal} constructs prototypes of different cardinalities for query and support videos, and calculate similarity by matching all prototype pairs. However, since TRX can only match prototypes with the same cardinality (\eg, three-frame prototypes matches with three-frame prototypes), it is hard to align the actions with different evolving speeds (\eg, one takes 2 frames while the other takes 4 frames). Also, the exhaustively matching of all pairs is computationally expensive.

We summarize three main differences compared with previous works: (1) We encode spatiotemporal object information to form more robust prototypes. (2) We generate a compound of global and focused prototypes to represent the actions from diverse perspectives. (3) The two groups of prototypes are efficiently matched to robustly compute video similarity.

\textbf{Transformers}~\cite{vaswani2017attention} have recently acquired remarkable achievements in computer vision~\cite{carion2020end,liu2021swin,deng2021transvg,doersch2020crosstransformers,zhang2021temporal,yang2021focal,yang2021stacked,huang2021leveraging}. FEAT~\cite{ye2020few} represents early work that applies transformer in the few-shot learning task, and TRX~\cite{perrett2021temporal} first introduces Transformer~\cite{doersch2020crosstransformers} into the few-shot action recognition task. Different from TRX, we apply a Transformer encoder-decoder to generate compound prototypes, and show that this is more effective in the few-shot action recognition scenario.

\section{Method}

\subsection{Problem Setting}
In few-shot action recognition, a model aims to recognize an unlabeled video (query) into one of the target categories each with a limited number of labeled examples (support set)~\cite{cao2020few,perrett2021temporal}. We follow the common practice~\cite{vinyals2016matching,finn2017model} to use episodic training, where in each episode a $C$-way $K$-shot problem is sampled: the support set $\mathcal{S}=\{\bm{X}^{j}_s\}_{j=1}^{C\times K}$ is composed of $C\times K$ labeled videos from $C$ different classes where each class contains $K$ samples. The query set contains $N$ unlabeled videos $\mathcal{Q} = \{\bm{X}_q^i\}_{i=1}^N$. The goal is to classify each video in the query set as one of the $C$ classes.

\subsection{Proposed Method}
\begin{figure}
    \centering
    \includegraphics[width=\linewidth]{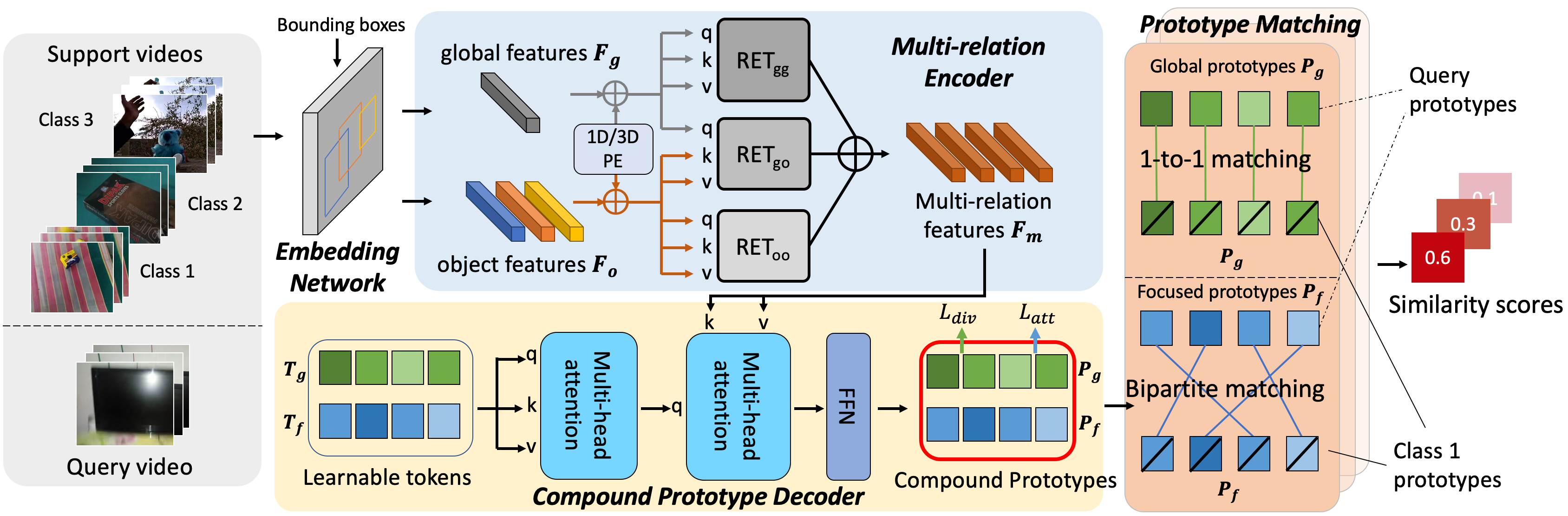}
    \caption{Illustration of our proposed method on a 3-way 1-shot problem. First, the videos are processed by an embedding network to acquire global (frame-level) features $\bm{F}_g$ and object features $\bm{F}_o$. Features $\bm{F}_g$ and $\bm{F}_o$ are equipped with 1D and 3D positional encoding (PE), respectively, and then used by a multi-relation encoder (\cref{sec:encoder}) to encode global-global, global-object and object-object information into a multi-relation feature $\bm{F}_m$. Using $\bm{F}_m$, a Transformer-based compound prototype decoder transforms the learnable tokens $\bm{T}_g, \bm{T}_f$ into compound prototypes that represent the input video (\cref{sec:decoder}). The compound prototypes is consisted of several global prototypes $\bm{P}_g$ (green squares) and focused prototypes $\bm{P}_f$ (blue squares). They are applied with different loss and different matching strategies, so that each global prototype captures a certain aspect of the action summarized from the whole video, and each focused prototype focuses on a specific temporal location of the video. The final similarity score is calculated as the average similarity of all matched prototype pairs between support and query videos. }
    \label{fig:model}
\end{figure}

We propose a new method for few-shot action recognition by generating and matching compound prototypes between query and support videos. As shown in \cref{fig:model}, given the video input, a feature embedding network first extracts global (frame-wise) features $\bm{F}_g$. To better model the actions involving multiple objects, we obtain object bounding boxes by a pretrained object detector~\cite{he2017mask}, and extract object features $\bm{F}_o$ using the same embedding network. Then a multi-relation encoder uses $\bm{F}_g$ and $\bm{F}_o$ to output multi-relation features $\bm{F}_m$ containing spatiotemporal global-object relations. Then a compound prototype decoder generates global prototypes $\bm{P}_g$ and focused prototypes $\bm{P}_f$ for each video. 
During similarity calculation, we use fixed 1-to-1 matching on the global prototypes and bipartite matching on the focused prototypes, encouraging the similarity to be computed robustly from diverse perspectives. We introduce the details of each component below.

\subsubsection{Feature embedding}
For each input video $\bm{X} \in \mathcal{S} \cup \mathcal{Q}$, we sample $T$ frames following the sampling strategy of TSN~\cite{wang2016temporal}. We use an embedding network to acquire a global (frame-level) feature representation for each video $\bm{F}_g\in\mathbb{R}^{T\times d}$, where $d$ is the feature dimension. Additionally we extract object features via ROI-Align~\cite{he2017mask} using the predicted bounding boxes on each frame. We use only $B$ most confident boxes on each frame, forming object features $\bm{F}_o\in\mathbb{R}^{BT\times d}$. 

\subsubsection{Multi-relation encoder}
\label{sec:encoder}
To better generate prototypes that are discriminative for actions involving multiple objects, we propose to use a multi-relation encoder to encode the spatiotemporal information from $\bm{F}_g$ and $\bm{F}_o$. 
We specifically consider the following three relations: global-global (frame-wise) relation, global-object relation, object-object relation, and apply transformer~\cite{vaswani2017attention} as the base architecture to allow relation modeling across frames. As shown in \cref{fig:model}, the encoder consists of three relation encoding transformers (RETs). The global-global RET (RET$_{gg}$) and the object-object RET (RET$_{oo}$) are identical except the input. They take as input the global feature $\bm{F}_g$ and the object feature $\bm{F}_o$ respectively, and use the input to generate the query $\bm{Q}$, key $\bm{K}$ and value $\bm{V}$ vectors for the transformer:
\begin{gather}
    \bm{F}_{gg} = \mathit{RET}_{gg}(\bm{Q}=\bm{K}=\bm{V}=\bm{F}_g) \\
    \bm{F}_{oo} = \mathit{RET}_{oo}(\bm{Q}=\bm{K}=\bm{V}=\bm{F}_o),
\end{gather}
The global-object RET (RET$_{go}$) works differently, where it maps $\bm{F}_g$ as query vector, while $\bm{F}_o$ as key and value vectors:
\begin{equation}
\resizebox{!}{!}{
    $\bm{F}_{go} = \mathit{RET}_{go}(\bm{Q}=\bm{F}_g,\; \bm{K}=\bm{V}=\bm{F}_o)$.
    }
\end{equation}
The output size of each RET is the same as its input query vector. Thus, each of the $T$ frame would have $B+2$ features with dimension $d$. We concatenate the three outputs into a multi-relation feature $\bm{F}_m\in\mathbb{R}^{(B+2)T\times d}$. 

Positional encoding (PE) is shown to be effective in transformer-based architectures~\cite{vaswani2017attention,lu2021simpler,liu2021swin}. We also use PE but omit in the equations for simplicity. For $\bm{F}_{g}$ we use 1D PE to encode the temporal location of each frame, and for $\bm{F}_{o}$ we apply 3D PE, encoding both spatial and temporal location of each object.

\subsubsection{Compound prototype decoder}
\label{sec:decoder}

The compound prototype decoder also follows the transformer architecture~\cite{vaswani2017attention,sun2021lesion,cong2021spatial,carion2020end}, so that the prototypes can be generated by considering feature of all frames via self-attention. As shown in \cref{fig:model}, the input to the prototype decoder are two groups of learnable tokens $\bm{T}_g\in \mathbb{R}^{m_{g} \times d}$ and $\bm{T}_f\in \mathbb{R}^{m_{f} \times d}$. 
A multi-head attention layer first encodes the tokens into $\bm{\hat{T}}_g$ and $\bm{\hat{T}}_f$, then another multi-head attention layer transforms them into two groups of prototypes $\bm{P}_g=\{\bm{p}_{g,k}\}_{k=1}^{m_g}\in \mathbb{R}^{m_{g} \times d}$ and $\bm{P}_f=\{\bm{p}_{f,k}\}_{k=1}^{m_f}\in \mathbb{R}^{m_{f} \times d}$. For simplicity, we omit the subscripts $_{g,f}$ and all normalization layers, thus the equation can be written as:
\begin{equation}
    \bm{Q} = \bm{\hat{T}}\bm{W}_Q, \quad \bm{K} = \bm{F}_m\bm{W}_K, \quad \bm{V} = \bm{F}_m\bm{W}_V,
\end{equation}
where $\bm{W}_Q, \bm{W}_K, \bm{W}_V \in \mathbb{R}^{d \times d}$ are linear projection weights, then we have
\begin{equation}\label{eqa:self-att}
    \bm{A} = softmax(\frac{\bm{Q}\bm{K}^T}{\sqrt{d}}), \quad \bm{P} = FFN(\bm{A}\bm{V}),
\end{equation}
where $\bm{A} \in \mathbb{R}^{m\times(B+2)T}$ is the self-attention weights, and FFN denotes feed forward network. 

To encourage the prototypes to capture different aspects of the action, we apply constraints on the two groups of prototypes individually. For the global prototypes $\bm{P}_g$, we add a diversity loss to maximize their diversity:
\begin{equation}
    L_{div} = \sum_{i\neq j} sim(\bm{p}_{g,i},\; \bm{p}_{g,j}),
\end{equation}
where $sim$ denotes the cosine similarity function.

Since learning $\bm{P}_g$ to robustly represent each aspect of the action (\eg, the start of the action) is difficult even with annotation~\cite{xu2020g,zeng2019graph}. To increase the overall robustness, for the focused prototypes $\bm{P}_f$, we instead add regularization on the self-attention weight $\bm{A}_f$ so that different $\bm{p}_f$ can focus on different temporal locations of the video:
\begin{equation}
    L_{att} = \sum_{i\neq j} sim(\bm{\alpha}_{f,i},\; \bm{\alpha}_{f,j}).
\end{equation}
Here $\bm{\alpha}_{f,i} \in \mathbb{R}^{(B+2)T}$ denotes the $i$-th row in $\bm{A}_f$. 

\subsubsection{Compound Prototype Matching}
Cooperating with the compound prototypes, we use different matching strategies to calculate the overall similarity between two videos. As shown in \cref{fig:model}, for the global prototypes $\bm{P}_g$, we match the query and support prototypes in a 1-to-1 manner, \ie, the $i$-th prototype of the query video is always matched with the $i$-th prototype of the support video. 
To calculate the global prototypes' overall similarity score between video $a$ and video $b$, we average the similarity score of all the $m_g$ global prototypes:
\begin{equation}
s_g^{a,b} = \frac{1}{m_g}\sum_{k=1}^{m_g}sim(\bm{p}_{g,k}^a,\; \bm{p}_{g,k}^b).
\end{equation}
Thus, to maximize the similarity score of correct video pairs and minimize the similarity of incorrect video pairs during episodic training, each $\bm{p}_g$ will try to encode a certain aspect of the action, \eg, the start of the action. This phenomenon is supported by our experiments in \cref{sec:experiments}.

For the focused prototypes $\bm{P}_f$, we apply a bipartite matching-based similarity measure.
Since different actions may happen in different temporal positions in the videos, the bipartite matching enables the temporal alignment of actions, allowing the comparison between actions of different lengths and at different speeds.
Formally speaking, for $\bm{P}_f^a$ of video $a$ and $\bm{P}_f^b$ of video $b$, we find a bipartite matching between these two sets of prototypes by searching for the best permutation of $m_f$ elements with the highest cosine similarity using the Hungarian algorithm~\cite{kuhn1955hungarian}. Denote $\sigma$ as the best permutation, the similarity score of $\bm{P}_f^a$ and $\bm{P}_f^b$ is calculated by:
\begin{equation}
s_d^{i,j} = \frac{1}{m_f}\sum_{k=1}^{m_f} sim (\bm{p}_{f,k}^a,\; \bm{p}_{f,\sigma(k)}^b).
\end{equation}

Finally, the similarity score is computed by a weighted average of $s_g$ and $s_f$: $s^{a,b} = \lambda_1 s_g^{a,b} + \lambda_2 s_f^{a,b}$. During training, this similarity score is directly regarded as logits for the cross-entropy loss $L_{ce}$. The total loss function is a weighted sum of three losses: $L = w_1L_{ce} + w_2L_{div} + w_3L_{att}$. During inference, we assign the label of the query video as the label of the most similar video in the support set.

\section{Experiments}
\label{sec:experiments}
We conduct experiments on four public datasets. \textbf{Kinetics}~\cite{carreira2017quo} and Something-something V2 (\textbf{SSv2})~\cite{goyal2017something} are two most frequently used benchmarks for few-shot action recognition. These two datasets are both splitted as 64/12/24 classes for train/val/test. For \textbf{SSv2}, we use both the split from CMN~\cite{zhu2018compound} (\textbf{SSv2$^\circ$}) and the split from OTAM~\cite{cao2020few} (\textbf{SSv2$^\sharp$}). Recently, Zhang \etal~\cite{zhang2020few} proposed new splits for \textbf{HMDB}~\cite{kuehne2011hmdb} and \textbf{UCF}~\cite{soomro2012ucf101} datasets. We also conduct experiments on these two datasets using the split from \cite{zhang2020few}.

Since our method's performance is competitive in both standard 1-shot 5-way setting and 5-shot 5-way setting, in this section we only demonstrate 1-shot results and place 5 shot results in the supplementary due to page limitation. Following previous works, we report the average result of 10,000 test episodes in the experiments.

\subsubsection{Baselines}
We compare our method with recent works reporting state-of-the-art performance, including MatchNet~\cite{vinyals2016matching}, CMN~\cite{zhu2020label}, OTAM~\cite{cao2020few}, TRN~\cite{zhou2018temporal}, ARN~\cite{zhang2020few}, TRX~\cite{perrett2021temporal}, ITA-Net~\cite{zhang2021learning}. Following \cite{zhu2021few}, we also compare with the few-shot image classification model FEAT~\cite{ye2020few} which is also based on transformers. Specifically, since no previous works used object detector in few-shot action recognition, for a fair comparison with previous works, we conduct experiments in two settings: (1) we give baseline methods the same input (both $\bm{F}_g$ and $\bm{F}_o$) as our method and compare the performance. We denote the baselines as ``Baseline+" in this setting. (2) We discard the object detector in our method and use only $\bm{F}_g$ as input and $RET_{gg}$ as the encoder. We denote our method in this setting as ``Ours-".

To enable previous methods to take object features as input, we choose to compare with methods TRX+, FEAT+, and ITA-Net+ because they also use transformer-based architectures like our method, thus no modification on the model architecture is needed. 
For completeness, we also compare MatchNet+ and TRN+ without transformer architecture. For these two methods, we reshape the object features $\bm{F}_o$ to $\bm{F}_o' \in \mathbb{R}^{T\times Bd}$ as input. Since these two works are not designed to input the object features, we train an ensembled network, one with $\bm{F}_g$ as input and the other with $\bm{F}_o'$ as input. We use the public implementation of TRX and implement ITA-Net by ourselves. More details about baseline implementation can be found in the supplementary material. 
Recently, several works~\cite{wang2021semantic,cao2021few,zhu2021few} pretrain the backbone on the meta-training set and found this pretraining to be useful for few-shot action recognition. To compare with the majority of prior works, we do not follow this setting in our experiments.

\subsubsection{Implementation details} 
We use ResNet-50~\cite{he2016deep} pretrained on ImageNet~\cite{deng2009imagenet} as the backbone of embedding network, and a fixed Mask-RCNN~\cite{he2017mask} trained on COCO dataset~\cite{lin2014microsoft} is used as the bounding box extractor. We select $B=3$ most confident object bounding boxes per frame. The pre-processing steps follow OTAM~\cite{cao2020few} where we also sample $T=8$ frames with random crop during training and center crop during inference. The model and the backbone is optimized using SGD with an initial learning rate of 0.001 and decaying every 20 epochs by 0.1. The embedding network except its first BatchNorm layer is fine-tuned with 1/10 learning rate. The training is stopped when the loss on the validation set is greater than the average of the previous 5 epochs. Unless otherwise stated, we report result using $m_g=m_f=8$, $\lambda_1=\lambda_2=0.5$, $w_1=1, w_2=w_3=0.1$. 
\begin{table}
  \centering
  \caption{Results of 5-way 1-shot experiments on 5 dataset splits. Methods marked with * indicates results of our implementation with the original reported results shown in parenthesis. The bottom and upper block are results with and without using object features, respectively. For ITA-Net, result on SSv2$^\circ$ comes from Table 5 of \cite{zhang2021learning}.}
  \label{tab:comparison}
  \scalebox{0.99}{
  \begin{tabular}{@{}lccccc@{}}
    \toprule
    Method & SSv2$^\circ$ & SSv2$^\sharp$ & Kinetics & HMDB & UCF \\
    \midrule
    MatchNet$^{*}$~\cite{vinyals2016matching}~~ & ~34.9 (31.3)~ & 35.1 & 54.6 (53.3) & 50.1 & 70.3 \\
    CMN~\cite{zhu2020label} & 36.2 & - & 60.5 & - & - \\
    ARN~\cite{zhang2020few} & - & - & 63.7 & 45.5 & 66.3 \\
    OTAM~\cite{cao2020few} & - & 42.8 & 73.0 & - & - \\
    TRN$^{*}$~\cite{zhou2018temporal,cao2020few} & 35.9 & ~39.6 (38.6)~ & ~68.6 (68.4)~ & 52.3 & 76.3 \\
    FEAT$^{*}$~\cite{ye2020few} & 38.4 & 45.5 & 73.0 & 56.1 & 75.8 \\
    ITA-Net$^{*}$~\cite{zhang2021learning} & 38.4 (38.6) & 46.1 & 72.6 & 56.5 & 76.0 \\
    TRX$^{*}$~\cite{perrett2021temporal} & 37.1 (36.0) & 41.5 (42.0) & 64.6 (63.6) & 54.4 & \textbf{77.7} \\
    Ours- & \textbf{38.9} & \textbf{49.3} & \textbf{73.3} & \textbf{60.1} & 71.4 \\
    \midrule
    MatchNet+$^{*}$ & 35.6 & 36.5 & 57.6 & 52.8 & 72.8 \\
    TRN+$^{*}$ & 38.1 & 41.3 & 71.4 & 55.3 & 79.7 \\
    FEAT+$^{*}$ & 43.1 & 46.6 & 73.9 & 61.5 & 79.7 \\
    ITA-Net+$^{*}$ & 43.7 & 48.9 & 74.4 & 61.6 & 79.5 \\
    TRX+$^{*}$ & 39.4 & 44.2 & 71.8 & 60.1 & \textbf{81.2} \\
    Ours & \textbf{57.1} & \textbf{59.6} & \textbf{81.0} & \textbf{80.3} & 79.0 \\
    \bottomrule
  \end{tabular}
  }
\end{table}

\subsection{Results}
\label{sec:results}
\cref{tab:comparison} shows result comparison with baseline methods. In the upper block, our model slightly outperforms previous works even without using the object detector on 4 of the 5 dataset/splits. This suggests that compared with other prototype generation methods, our proposed approach to generate and match the compound prototypes enables better similarity measurement for few-shot action recognition.
From the comparison of baseline methods with their ``+" variants, we can see that these methods cannot fully exploit the information brought by the object features. When object features are added as input, our full model significantly outperforms previous methods that use the same input. Compared with the performance of our method in the upper block, the object features significantly stimulates the potential of our proposed compound prototype matching scheme, by improving the accuracy on the SSv2 dataset for over 10\%, the Kinetics dataset for 7.7\% and the HMDB dataset for over 20\%. 
In the ablation study, we show even better performance can be achieved by carefully adjusting the number of prototypes $m_g$ and $m_f$. 
\begin{table}[]

\parbox{.5\linewidth}{
\centering
\caption{Results comparison of different methods when using/not using object information, and using/not using our proposed encoder. }
\label{tab:encoder}
\scalebox{0.85}{

\begin{tabular}{@{}ccccc@{}}
\toprule
Encoding                              & Prototype & ~SSv2$^\circ$ & SSv2$^\sharp$ & Kinetics \\ \midrule
\multirow{4}{*}{\begin{tabular}[c]{@{}c@{}}Backbone only\\ (w/o. object \\ features)\end{tabular}} & None & 34.9                  & 35.1                  & 54.6                  \\
                                      & ITA-Net  & \textbf{38.4}                  & 45.5                  & \textbf{72.6}                  \\
                                      & TRX      & 37.1                  & 41.5                  & 64.6                  \\
                                      & Ours     & 37.9                  & \textbf{48.5}                  & 72.5                  \\ \midrule
\multirow{4}{*}{\begin{tabular}[c]{@{}c@{}}Our encoder\\ (w/o. object\\ features)\end{tabular}}            & None & 35.3                  & 37.2                 & 59.4                 \\
                                      & ITA-Net  & 38.6                  & 46.5                  & 73.0                  \\
                                      & TRX      & 38.2                 & 44.4                  & 68.9                  \\
                                      & Ours     & \textbf{38.9}                  & \textbf{49.3}                  & \textbf{73.3}                  \\ \midrule

\multirow{4}{*}{\begin{tabular}[c]{@{}c@{}}Concat\\ (w. object\\ features)\end{tabular}}        & None & 35.6                  & 36.5                 & 57.6                  \\
                                      & ITA-Net+  & \textbf{43.7}                  & 48.9                  & 74.4                  \\
                                      & TRX+      & 39.4                  & 44.2                  & 71.8                  \\
                                      & Ours     & 42.1                  & \textbf{49.1}                  & \textbf{79.2}                  \\ \midrule
\multirow{4}{*}{\begin{tabular}[c]{@{}c@{}}Our encoder\\ (w. object\\ features)\end{tabular}}                 & None & 39.1                  & 42.5                  & 58.4                  \\
                                      & ~ITA-Net+  & 44.0                    & 50.7                  & 73.3                  \\
                                      & TRX+      & 46.3                  & 48.4                  & 73.4                  \\
                                      & Ours     & \textbf{57.1}                  & \textbf{59.6}                  & \textbf{81.0}                  \\ \bottomrule
\end{tabular}
}
}
\hfill
\parbox{0.45\linewidth}{
\centering
\caption{Results comparison of our model using different encoding relations with different number of global prototypes and focused prototypes.}
\label{tab:global}
\scalebox{0.9}{
\begin{tabular}{@{}ccccccc@{}}
\toprule
Setting & $\bm{F}_{gg}$ & $\bm{F}_{go}$ & $\bm{F}_{oo}$ & SSv2$^\sharp$ & Kinetics \\
\midrule
\multirow{7}{*}{\makecell[c]{$m_g=8$,\\$m_f=8$}} 
& \checkmark & $\times$ & $\times$ & 49.3 & 73.3 \cr
& $\times$ & \checkmark & $\times$ & 47.3 & 69.1 \cr
& $\times$ & $\times$ & \checkmark  & 55.5 & 73.6 \cr
& \checkmark & \checkmark & $\times$ & 49.6 & 75.0 \cr
& \checkmark & $\times$ & \checkmark & 57.7 & 79.7 \cr
& $\times$ & \checkmark & \checkmark & 57.8 & 77.1 \cr
& \checkmark & \checkmark & \checkmark & \textbf{59.6} & \textbf{81.0} \cr
\midrule
\multirow{4}{*}{\makecell[c]{$m_g=16$,\\$m_f=0$}} 
& \checkmark & $\times$ & $\times$ & 38.9 & 65.5 \cr
& $\times$ & \checkmark & $\times$ & 38.8 & 65.2 \cr
& $\times$ & $\times$ & \checkmark  & 38.5 & 64.0 \cr
& \checkmark & \checkmark & \checkmark & \textbf{39.3} & \textbf{66.1} \cr
\midrule
\multirow{4}{*}{\makecell[c]{$m_g=0$,\\$m_f=16$}} 
& \checkmark & $\times$ & $\times$ & 41.9 & 72.3 \cr
& $\times$ & \checkmark & $\times$ & 34.4 & 73.7 \cr
& $\times$ & $\times$ & \checkmark  & 29.3 & 61.0 \cr
& \checkmark & \checkmark & \checkmark & \textbf{44.0} & \textbf{76.7} \cr
\bottomrule
\end{tabular}
}
}
\end{table}

Our method does not achieve state-of-the-art results on the UCF dataset. One reason is that classes in the UCF dataset can be easily distinguished only from appearance. This causes our multi-relation encoder to overfit. If we remove this encoder and directly use concatenation of object and global features as input to the compound prototype decoder, our method can achieve the state-of-the-art accuracy of 87.7. This reveals one limitation of our method, \ie, easy to overfit on simple datasets.

\subsection{Ablation Study}
\label{sec:ablation}
\subsubsection{Effect of object features}
While there exist previous studies that leverage additional information for few-shot action recognition~\cite{fu2020depth}, no work has investigated the use of object features as in our method. One may argue that the performance improvement of our method only comes from the use of object features. However, in \cref{tab:comparison} we find that performance gain of previous methods is limited if object features are additionally used as input. Here, we show that a boost in performance only happens when object features, our multi-relation encoder, and our decoder are used together.

We test the performance of multiple methods with and without using object information, and also with and without using our multi-relation encoder.
In \cref{tab:encoder}, the first block uses neither object feature nor our multi-relation encoder, and the second block uses our encoder but with only the global-global relation $RET_{gg}$.
From the comparison between these two blocks, we can see that our multi-relation encoder can improve the performance of all methods, but not significantly.
In the third block of \cref{tab:encoder}, we concatenate each frame-wise feature with its corresponding object features as input. The comparison between this block and the first block shows the improvement brought by object features: around 2 $\sim$ 5 percent on SSv2 dataset, and 2 $\sim$ 6 percent on Kinetics dataset.
Finally, in the fourth block of \cref{tab:encoder}, both object features and our multi-relation encoder are used. Comparing this block with the second and third blocks, all methods get more improvement. This shows that using our multi-relation encoder to consider multiple relations across frames can better leverage the information brought by the object features. Among all methods in the fourth block of \cref{tab:encoder}, our method enjoys the most significant performance gain. This indicates that while the object features bring additional information, our method can best leverage this information to help the few-shot action recognition task.

\subsubsection{Impact of multi-relation feature encoding}
To see how does the multi-relation feature encoding contribute to the performance, we conduct ablation study of our method considering only subsets of $\{\bm{F}_{gg}, \bm{F}_{go}, \bm{F}_{oo}\}$. We also vary the number of global prototypes $m_g$ and focused prototypes $m_f$ to see the influence of feature encoding on each group of prototypes.

Results can be seen in \cref{tab:global}. From the experiments with $m_g=m_f=8$, the SSv2$^\sharp$ dataset gets much improvement from the use of object-object feature $\bm{F}_{oo}$, while the Kinetics dataset benefits more from the global-global feature $\bm{F}_{gg}$. This is reasonable since SSv2 dataset includes more actions with multiple objects. From the experiments with $m_g=16$, the global prototypes seem to work equally well using the three encoded features on both datasets. The experiments with $m_f=16$ suggest that focused prototypes work better with global-global relations. When using all three features (last row of each block), the comparison between different choices of $m_g$ and $m_f$ indicates that the two groups of prototypes capture complementary aspects of the action, since the performances got significantly improved when two groups of prototypes both present. A more detailed figure showing the accuracy difference of each action class can be found in the supplementary material.

\subsection{Analysis of compound prototypes}
\label{sec:ablation-prototyps}
The core of our proposed method is the generation and matching of compound prototypes. In this section we conduct extensive experiments to get a more comprehensive understanding of the two groups of prototypes. 

\begin{figure*}[t]
    \centering
    \includegraphics[width=\linewidth]{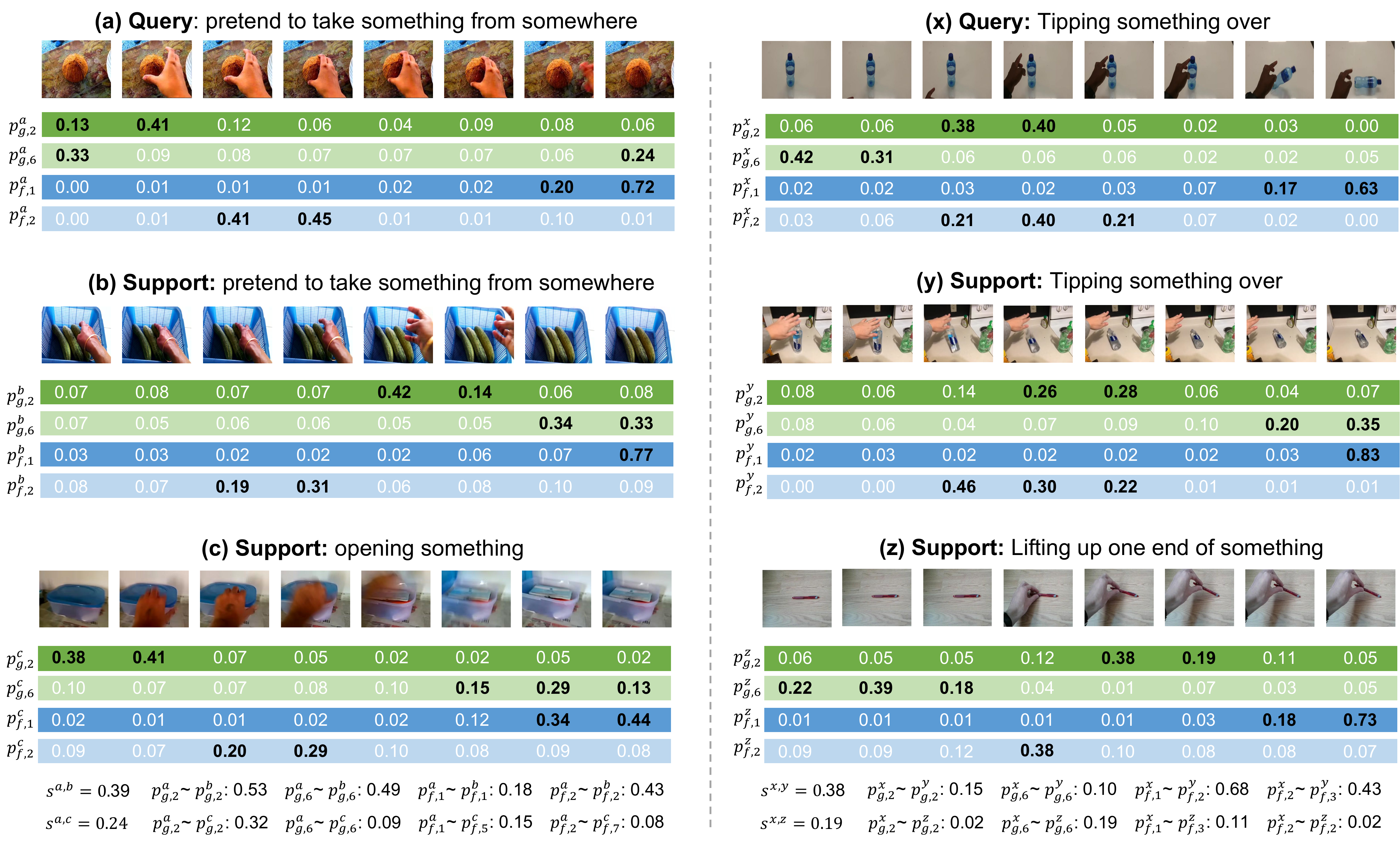}
    \caption{Visualization of the self-attention weight of two \textcolor{ForestGreen}{global prototypes} and two \textcolor{blue}{focused prototypes} on each timestamp of the input. Attention weights higher than average (0.125) are marked in black. We can see the global prototypes capture a certain aspect of the action in the video, regardless of temporal location: $\bm{p}_{g,2}$ - the start of the action; $\bm{p}_{g,6}$ - the frames without hand. Meanwhile, the focused prototypes mainly attend on fixed timestamps of the video: $\bm{p}_{f,1}$ - the end of the video; $\bm{p}_{f,2}$ - the middle part of the video. Example to the left comes from the SSv2$^\circ$ dataset and the example to the right is from SSv2$^\sharp$.
    Video similarity scores $s$ and similarity scores of matched prototypes $\bm{p}_*\sim \bm{p}_*$ are shown at the bottom. }
    \label{fig:visualize}
\end{figure*}

\subsubsection{What aspect of the action does each prototype capture?} 
We address this question by investigating the self-attention operation that generates the prototypes. From Equation~\ref{eqa:self-att}, the self-attention weight $\bm{\alpha}\in \mathbb{R}^{(B+2)\times T}$ on each frame represents from which part of the video does each prototype gather its information. To better understand the prototypes, we visualize this attention in \cref{fig:visualize} using two 1-shot 2-way examples. In the visualization we average the $B+2$ attention weights in each frame, forming $\bm{\tilde{\alpha}} \in \mathbb{R}^{T}$, and show this averaged value on each of the $T=8$ frames. For clarity we only show 2 \textcolor{ForestGreen}{global prototypes} and 2 \textcolor{blue}{focused prototypes} in each video. We also show the video similarity scores and the similarity of matched prototypes at the bottom of the figure.

The visualization is shown in \cref{fig:visualize}. We first analyze the attention of each prototype. In all videos, the global prototype $\bm{p}_{g,2}$ have high attention weights on the start of the \textit{action} (not the start of the \textit{video}), and $\bm{p}_{g,6}$ pays more attention to the frames that contain appearance change compared with other frames (no hand existence). This is expected since $L_{div}$ forces each global prototype to be different, while the 1-to-1 matching encourages each global prototype to focus on similar aspects so that correct video similarity can be predicted.
For the focused prototypes, $\bm{p}_{f,1}$ usually gives high attention to the last few frames, and $\bm{p}_{f,2}$ pays more attention on the middle frames. This is also expected since $L_{att}$ refrains the focused prototypes to attend on similar temporal locations, and bipartite matching allows similar actions to be matched even when they are at different temporal locations of the videos. A similar phenomenon exists in the object detection task~\cite{carion2020end}, where each object query focuses on detecting objects in a specific spatial location of the image.

In the left example, both the prototype pairs $<\bm{p}_{g,2}^a,\bm{p}_{g,2}^b>$ and $<\bm{p}_{g,6}^a,\bm{p}_{g,6}^b>$ have high similarity scores (0.53 and 0.49 shown at the bottom of the left example). This indicates that video $a$ and $b$ have similar starts, and the intra-video appearance change is also similar. Thus the query action is correctly classified as ``Pretending to take something from somewhere". 
In the right example, we can see the effectiveness of the focused prototypes.  By Hungarian matching, $\bm{p}_{f,1}^x$ is matched with $\bm{p}_{f,2}^y$. Since they both encode the frames where the hands just tip the objects over, these two prototypes give high similarities, enabling the query action of ``Tipping something over" to be correctly recognized.

\begin{figure}[t]
    \centering
    \parbox{0.48\linewidth}{
    \includegraphics[width=\linewidth]{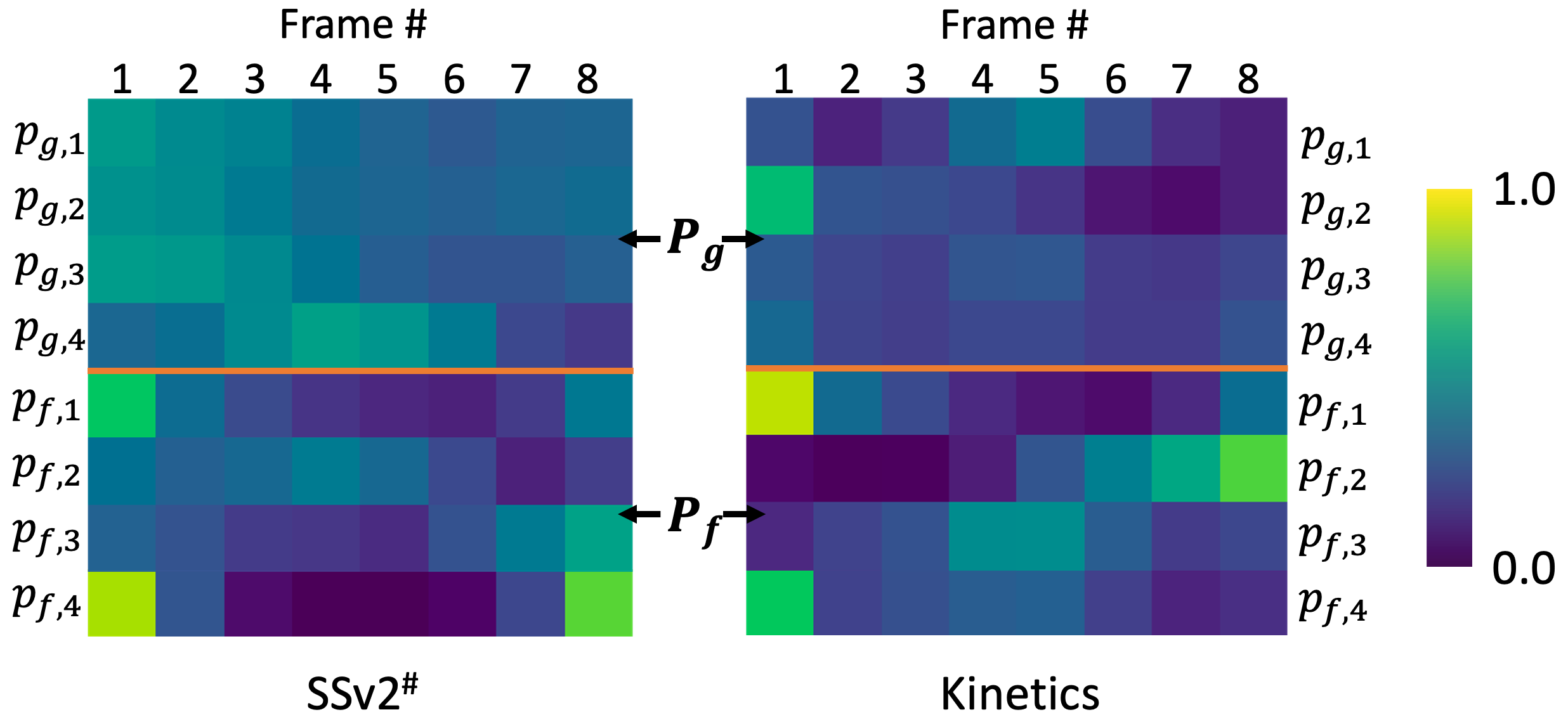}
    \caption{Visualization of self-attention weights of $\bm{P}_s$ and $\bm{P}_d$ for all samples on the test set of SSv2$^\sharp$ and Kinetics datasets.}
    \label{fig:weight}
   }
   \hfill
   \parbox{0.48\linewidth}{
    \includegraphics[width=\linewidth]{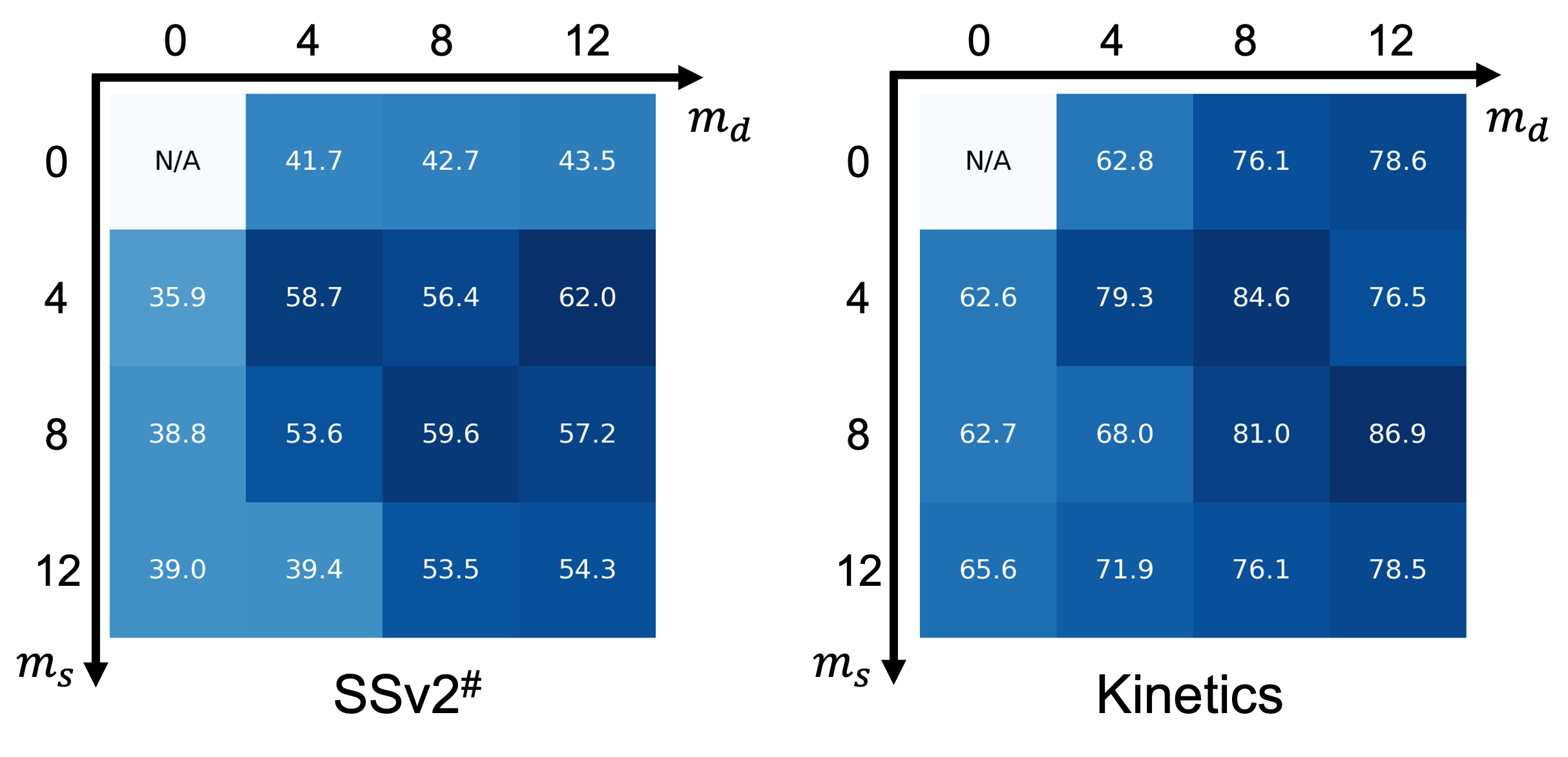}
    \caption{Performance on the SSv2$^\sharp$ and Kinetics datasets when changing the number of global/focused prototypes.}
    \label{fig:param}
    }
\end{figure}

A statistical analysis of self-attention weights can be found in \cref{fig:weight} showing the average response of the first 4 global prototypes and the first 4 focused prototypes on all videos of the test set. As a result of the loss functions $L_{div},L_{att}$ and the matching strategies, on both SSv2$^\sharp$ and Kinetics datasets, $\bm{P}_g$ (first 4 rows) have a more uniform attention distribution, while $\bm{P}_f$ have obvious temporal regions of focus. The diversity of the prototypes ensures a robust representation of the videos, thus similarity between videos can be better computed during the few-shot learning process.

\subsubsection{How much does each group of prototype contribute?} 
\begin{figure}[t]
    \centering
    \includegraphics[width=0.85\linewidth]{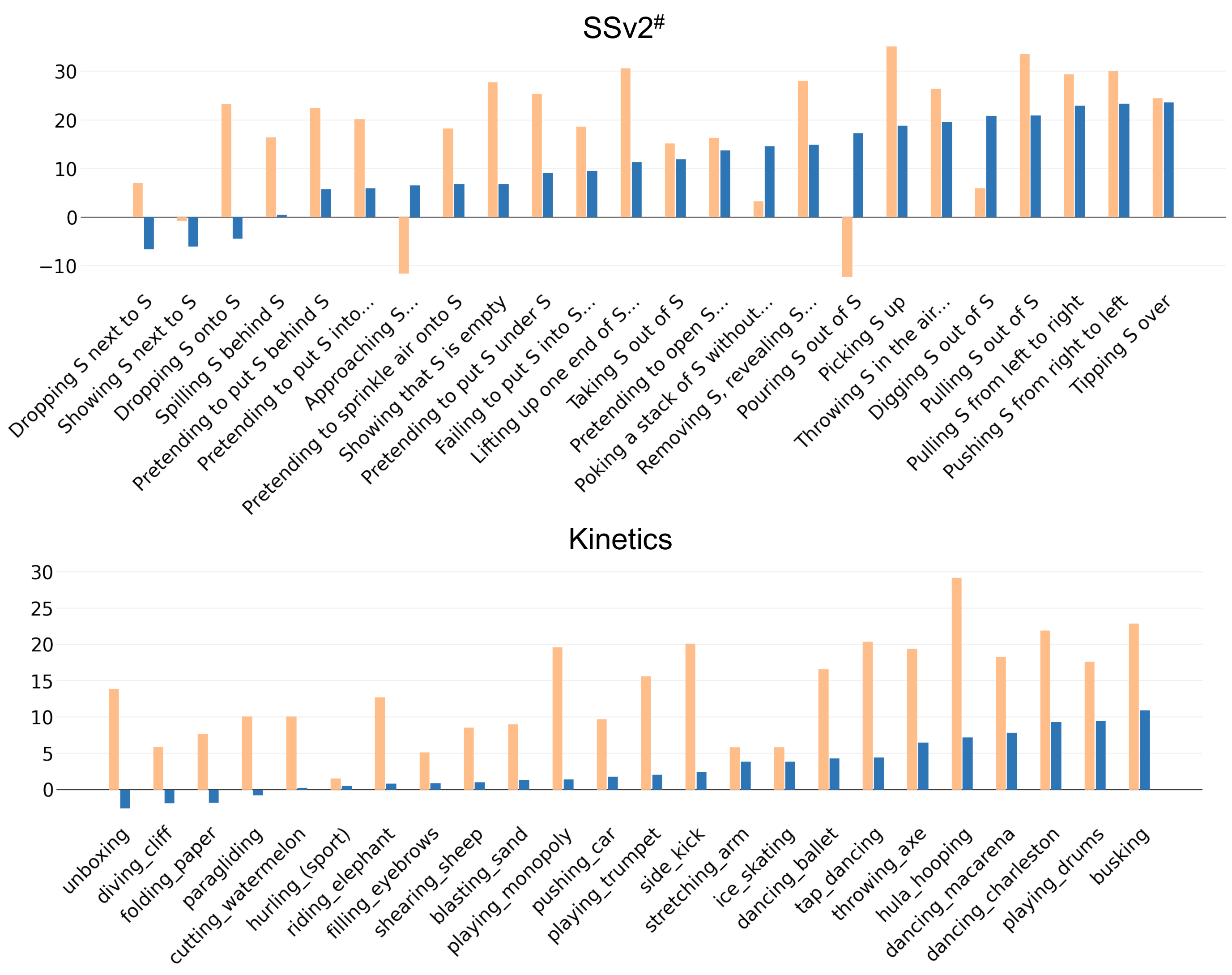}
    \caption{Class accuracy improvement when our method uses $m_g=m_f=8$ prototypes compared to: orange bars: $m_g=16$, $m_f=0$; blue bars: $m_g=0$, $m_f=16$. S stands for the abbreviation of ``something".}
    \label{fig:class}
\end{figure}
To find the answer, we test our method using different numbers of prototypes ($m_g$ and $m_f$) and show the results in \cref{fig:param}. Although the best combination of $m_g$ and $m_f$ are different for each dataset, the performance gets better when the number of prototypes increases, and after a certain threshold, the result saturates because of the overfitting on the training data. Best results on both SSv2$^\sharp$ (62.0) and Kinetics (86.9) datasets are achieved when $m_f$ is larger than $m_g$. Although $m_g=m_f=8$ is not the optimal setting, we apply this setting in \cref{sec:results} and \cref{sec:ablation} since it is the most stable setting on all datasets. A method to automatically choose the number of prototypes is left for our future work.

Also, we show the class accuracy improvement when our method uses both groups of prototypes compared with our method using only one group of prototype. In \cref{fig:class}, orange bars denote the accuracy difference between the $m_g=m_f=8$ setting and the $m_g=16, m_f=0$ setting, which indicates the performance gain by introducing the focused prototypes. Blue bars, on the other hand, show the accuracy improvement brought by the global prototypes. 
We can see on the SSv2$^\sharp$ dataset that when combining the two groups of prototypes, some hard classes like ``pulling S out of S", ``pulling S from left to right" and ``pushing S from right to left" can be better distinguished. 
From the results of the two datasets, we observe the focused prototypes are more effective in the Kinetics dataset. This is because the Kinetics dataset focuses more on appearance, which can be better captured and compared by the focused prototypes.

\subsubsection{Will wrong bipartite matching destroy the temporal ordering of actions?}
Although we observe great performance gain brought by $\bm{P}_f$ in \cref{tab:global} and \cref{fig:class}, the bipartite matching will unavoidably produce some wrongly matched prototype pairs when creating the correct matchings. 
% In our experiment, we found that these wrong matchings are essential in the training process. 
Our additional experiments in the supplementary show that filtering the matched prototypes with low similarity negatively affects the convergence of the model. One reason is that positional encoding implicitly encodes the temporal ordering of the frames within each prototype. During training, the similarity scores of all the wrong matching pairs are learned to be small and so that the final similarity score can be dominated by the similarity score of the correctly matched prototype pairs.

\section{Conclusion}
In this work, we introduce a novel method for few-shot action recognition by generating global and focused prototypes and compare video similarity based on the prototypes. When generating the prototypes, we encode spatiotemporal object relations to address the actions that involve multiple objects. The two groups of prototypes are encouraged to capture different aspects of the input by different loss functions and matching strategies.
% Our method achieves state-of-the-art performance on multiple datasets. 
In our future work, we will explore a more flexible prototype matching strategy that can avoid the mismatch in the bipartite matching.

\noindent\textbf{Acknowledgement~} This work is supported by JSPS KAKENHI Grant Number JP22K17905, JP20H04205 and JST AIP Acceleration Research Grant Number JPMJCR20U1.
\bibliographystyle{splncs04}
\bibliography{egbib}

\begin{thebibliography}{10}
\providecommand{\url}[1]{\texttt{#1}}
\providecommand{\urlprefix}{URL }
\providecommand{\doi}[1]{https://doi.org/#1}

\bibitem{andrychowicz2016learning}
Andrychowicz, M., Denil, M., Gomez, S., Hoffman, M.W., Pfau, D., Schaul, T.,
  Shillingford, B., De~Freitas, N.: Learning to learn by gradient descent by
  gradient descent. In: NeurIPS (2016)

\bibitem{antoniou2018train}
Antoniou, A., Edwards, H., Storkey, A.: How to train your maml. ICML  (2019)

\bibitem{bishay2019tarn}
Bishay, M., Zoumpourlis, G., Patras, I.: Tarn: Temporal attentive relation
  network for few-shot and zero-shot action recognition. BMVC  (2019)

\bibitem{cao2021few}
Cao, C., Li, Y., Lv, Q., Wang, P., Zhang, Y.: Few-shot action recognition with
  implicit temporal alignment and pair similarity optimization. CVIU  (2021)

\bibitem{cao2020few}
Cao, K., Ji, J., Cao, Z., Chang, C.Y., Niebles, J.C.: Few-shot video
  classification via temporal alignment. In: CVPR (2020)

\bibitem{carion2020end}
Carion, N., Massa, F., Synnaeve, G., Usunier, N., Kirillov, A., Zagoruyko, S.:
  End-to-end object detection with transformers. In: ECCV (2020)

\bibitem{carreira2017quo}
Carreira, J., Zisserman, A.: Quo vadis, action recognition? a new model and the
  kinetics dataset. In: CVPR (2017)

\bibitem{chang2019d3tw}
Chang, C.Y., Huang, D.A., Sui, Y., Fei-Fei, L., Niebles, J.C.: D3tw:
  Discriminative differentiable dynamic time warping for weakly supervised
  action alignment and segmentation. In: CVPR (2019)

\bibitem{chao2018rethinking}
Chao, Y.W., Vijayanarasimhan, S., Seybold, B., Ross, D.A., Deng, J.,
  Sukthankar, R.: Rethinking the faster r-cnn architecture for temporal action
  localization. In: CVPR (2018)

\bibitem{chowdhury2021few}
Chowdhury, A., Jiang, M., Chaudhuri, S., Jermaine, C.: Few-shot image
  classification: Just use a library of pre-trained feature extractors and a
  simple classifier. In: ICCV (2021)

\bibitem{cong2021spatial}
Cong, Y., Liao, W., Ackermann, H., Rosenhahn, B., Yang, M.Y.: Spatial-temporal
  transformer for dynamic scene graph generation. In: ICCV (2021)

\bibitem{deng2009imagenet}
Deng, J., Dong, W., Socher, R., Li, L.J., Li, K., Fei-Fei, L.: Imagenet: A
  large-scale hierarchical image database. In: CVPR (2009)

\bibitem{deng2021transvg}
Deng, J., Yang, Z., Chen, T., Zhou, W., Li, H.: Transvg: End-to-end visual
  grounding with transformers. ICCV  (2021)

\bibitem{dhillon2019baseline}
Dhillon, G.S., Chaudhari, P., Ravichandran, A., Soatto, S.: A baseline for
  few-shot image classification. ICLR  (2019)

\bibitem{doersch2020crosstransformers}
Doersch, C., Gupta, A., Zisserman, A.: Crosstransformers: spatially-aware
  few-shot transfer. NeurIPS  (2020)

\bibitem{fan2020few}
Fan, Q., Zhuo, W., Tang, C.K., Tai, Y.W.: Few-shot object detection with
  attention-rpn and multi-relation detector. In: CVPR (2020)

\bibitem{finn2017model}
Finn, C., Abbeel, P., Levine, S.: Model-agnostic meta-learning for fast
  adaptation of deep networks. In: International Conference on Machine
  Learning. PMLR (2017)

\bibitem{fu2020depth}
Fu, Y., Zhang, L., Wang, J., Fu, Y., Jiang, Y.G.: Depth guided adaptive
  meta-fusion network for few-shot video recognition. In: ACM MM (2020)

\bibitem{goyal2017something}
Goyal, R., Ebrahimi~Kahou, S., Michalski, V., Materzynska, J., Westphal, S.,
  et~al.: The" something something" video database for learning and evaluating
  visual common sense. In: ICCV (2017)

\bibitem{grauman2021ego4d}
Grauman, K., Westbury, A., Byrne, E., et~al.: Ego4d: Around the world in 3,000
  hours of egocentric video. arXiv preprint arXiv:2110.07058  (2021)

\bibitem{gui2018few}
Gui, L.Y., Wang, Y.X., Ramanan, D., Moura, J.M.: Few-shot human motion
  prediction via meta-learning. In: ECCV (2018)

\bibitem{he2017mask}
He, K., Gkioxari, G., Doll{\'a}r, P., Girshick, R.: Mask r-cnn. In: ICCV (2017)

\bibitem{he2016deep}
He, K., Zhang, X., Ren, S., Sun, J.: Deep residual learning for image
  recognition. In: CVPR (2016)

\bibitem{hong2021video}
Hong, J., Fisher, M., Gharbi, M., Fatahalian, K.: Video pose distillation for
  few-shot, fine-grained sports action recognition. In: ICCV (2021)

\bibitem{huang2018predicting}
Huang, Y., Cai, M., Li, Z., Sato, Y.: Predicting gaze in egocentric video by
  learning task-dependent attention transition. In: ECCV (2018)

\bibitem{huang2021leveraging}
Huang, Y., Li, X., Yang, L., Gu, L., Zhu, Y., Seo, H., Meng, Q., Harada, T.,
  Sato, Y.: Leveraging human selective attention for medical image analysis
  with limited training data. BMVC  (2021)

\bibitem{huang2020improving}
Huang, Y., Sugano, Y., Sato, Y.: Improving action segmentation via graph-based
  temporal reasoning. In: Proceedings of the IEEE/CVF conference on computer
  vision and pattern recognition. pp. 14024--14034 (2020)

\bibitem{kang2019few}
Kang, B., Liu, Z., Wang, X., Yu, F., Feng, J., Darrell, T.: Few-shot object
  detection via feature reweighting. In: ICCV (2019)

\bibitem{kang2021relational}
Kang, D., Kwon, H., Min, J., Cho, M.: Relational embedding for few-shot
  classification. In: ICCV (2021)

\bibitem{kliper2011one}
Kliper-Gross, O., Hassner, T., Wolf, L.: One shot similarity metric learning
  for action recognition. In: SIMBAD (2011)

\bibitem{koch2015siamese}
Koch, G., Zemel, R., Salakhutdinov, R., et~al.: Siamese neural networks for
  one-shot image recognition. In: ICML (2015)

\bibitem{kuehne2011hmdb}
Kuehne, H., Jhuang, H., Garrote, E., Poggio, T., Serre, T.: Hmdb: a large video
  database for human motion recognition. In: ICCV (2011)

\bibitem{kuhn1955hungarian}
Kuhn, H.W.: The hungarian method for the assignment problem. Naval research
  logistics quarterly  (1955)

\bibitem{kumar2019protogan}
Kumar~Dwivedi, S., Gupta, V., Mitra, R., Ahmed, S., Jain, A.: Protogan: Towards
  few shot learning for action recognition. In: CVPRW (2019)

\bibitem{li2019finding}
Li, H., Eigen, D., Dodge, S., Zeiler, M., Wang, X.: Finding task-relevant
  features for few-shot learning by category traversal. In: CVPR (2019)

\bibitem{li2021ta2n}
Li, S., Liu, H., Qian, R., Li, Y., See, J., Fei, M., Yu, X., Lin, W.: Ta2n:
  Two-stage action alignment network for few-shot action recognition. arXiv
  preprint arXiv:2107.04782  (2021)

\bibitem{lin2014microsoft}
Lin, T.Y., Maire, M., Belongie, S., Hays, J., Perona, P., Ramanan, D.,
  Doll{\'a}r, P., Zitnick, C.L.: Microsoft coco: Common objects in context. In:
  ECCV (2014)

\bibitem{liu2020crnet}
Liu, W., Zhang, C., Lin, G., Liu, F.: Crnet: Cross-reference networks for
  few-shot segmentation. In: CVPR (2020)

\bibitem{liu2020part}
Liu, Y., Zhang, X., Zhang, S., He, X.: Part-aware prototype network for
  few-shot semantic segmentation. In: ECCV (2020)

\bibitem{liu2021swin}
Liu, Z., Lin, Y., Cao, Y., Hu, H., Wei, Y., Zhang, Z., Lin, S., Guo, B.: Swin
  transformer: Hierarchical vision transformer using shifted windows. ICCV
  (2021)

\bibitem{lu2021simpler}
Lu, Z., He, S., Zhu, X., Zhang, L., Song, Y.Z., Xiang, T.: Simpler is better:
  Few-shot semantic segmentation with classifier weight transformer. In: ICCV
  (2021)

\bibitem{luo2020weakly}
Luo, Z., Guillory, D., Shi, B., Ke, W., Wan, F., Darrell, T., Xu, H.:
  Weakly-supervised action localization with expectation-maximization
  multi-instance learning. In: ECCV (2020)

\bibitem{ma2021weakly}
Ma, J., Gorti, S.K., Volkovs, M., Yu, G.: Weakly supervised action selection
  learning in video. In: CVPR (2021)

\bibitem{mishra2018generative}
Mishra, A., Verma, V.K., Reddy, M.S.K., Arulkumar, S., Rai, P., Mittal, A.: A
  generative approach to zero-shot and few-shot action recognition. In: WACV
  (2018)

\bibitem{patravali2021unsupervised}
Patravali, J., Mittal, G., Yu, Y., Li, F., Chen, M.: Unsupervised few-shot
  action recognition via action-appearance aligned meta-adaptation. In: ICCV
  (2021)

\bibitem{perrett2021temporal}
Perrett, T., Masullo, A., Burghardt, T., Mirmehdi, M., Damen, D.:
  Temporal-relational crosstransformers for few-shot action recognition. In:
  CVPR (2021)

\bibitem{qiao2018few}
Qiao, S., Liu, C., Shen, W., Yuille, A.L.: Few-shot image recognition by
  predicting parameters from activations. In: CVPR (2018)

\bibitem{ravi2016optimization}
Ravi, S., Larochelle, H.: Optimization as a model for few-shot learning. ICLR
  (2017)

\bibitem{snell2017prototypical}
Snell, J., Swersky, K., Zemel, R.S.: Prototypical networks for few-shot
  learning. NeurIPS  (2017)

\bibitem{soomro2012ucf101}
Soomro, K., Zamir, A.R., Shah, M.: Ucf101: A dataset of 101 human actions
  classes from videos in the wild. arXiv preprint arXiv:1212.0402  (2012)

\bibitem{sun2021lesion}
Sun, R., Li, Y., Zhang, T., Mao, Z., Wu, F., Zhang, Y.: Lesion-aware
  transformers for diabetic retinopathy grading. In: CVPR (2021)

\bibitem{thatipelli2021spatio}
Thatipelli, A., Narayan, S., Khan, S., Anwer, R.M., Khan, F.S., Ghanem, B.:
  Spatio-temporal relation modeling for few-shot action recognition. arXiv
  preprint arXiv:2112.05132  (2021)

\bibitem{vaswani2017attention}
Vaswani, A., Shazeer, N., Parmar, N., Uszkoreit, J., Jones, L., Gomez, A.N.,
  Kaiser, {\L}., Polosukhin, I.: Attention is all you need. In: NeurIPS (2017)

\bibitem{vinyals2016matching}
Vinyals, O., Blundell, C., Lillicrap, T., Wierstra, D., et~al.: Matching
  networks for one shot learning. NeurIPS  (2016)

\bibitem{wang2020few}
Wang, H., Zhang, X., Hu, Y., Yang, Y., Cao, X., Zhen, X.: Few-shot semantic
  segmentation with democratic attention networks. In: ECCV (2020)

\bibitem{wang2019panet}
Wang, K., Liew, J.H., Zou, Y., Zhou, D., Feng, J.: Panet: Few-shot image
  semantic segmentation with prototype alignment. In: ICCV (2019)

\bibitem{wang2016temporal}
Wang, L., Xiong, Y., Wang, Z., Qiao, Y., Lin, D., Tang, X., Van~Gool, L.:
  Temporal segment networks: Towards good practices for deep action
  recognition. In: ECCV (2016)

\bibitem{wang2021semantic}
Wang, X., Ye, W., Qi, Z., Zhao, X., Wang, G., Shan, Y., Wang, H.:
  Semantic-guided relation propagation network for few-shot action recognition.
  In: ACM MM (2021)

\bibitem{wang2020frustratingly}
Wang, X., Huang, T.E., Darrell, T., Gonzalez, J.E., Yu, F.: Frustratingly
  simple few-shot object detection. ICML  (2020)

\bibitem{wei2019piecewise}
Wei, X.S., Wang, P., Liu, L., Shen, C., Wu, J.: Piecewise classifier mappings:
  Learning fine-grained learners for novel categories with few examples. TIP
  (2019)

\bibitem{xian2020generalized}
Xian, Y., Korbar, B., Douze, M., Schiele, B., Akata, Z., Torresani, L.:
  Generalized many-way few-shot video classification. In: ECCV (2020)

\bibitem{xian2021generalized}
Xian, Y., Korbar, B., Douze, M., Torresani, L., Schiele, B., Akata, Z.:
  Generalized few-shot video classification with video retrieval and feature
  generation. TPAMI  (2021)

\bibitem{xu2018dense}
Xu, B., Ye, H., Zheng, Y., Wang, H., Luwang, T., Jiang, Y.G.: Dense dilated
  network for few shot action recognition. In: ICMR (2018)

\bibitem{xu2021learning}
Xu, C., Fu, Y., Liu, C., Wang, C., Li, J., Huang, F., Zhang, L., Xue, X.:
  Learning dynamic alignment via meta-filter for few-shot learning. In: CVPR
  (2021)

\bibitem{xu2020g}
Xu, M., Zhao, C., Rojas, D.S., Thabet, A., Ghanem, B.: G-tad: Sub-graph
  localization for temporal action detection. In: CVPR (2020)

\bibitem{yang2021focal}
Yang, J., Li, C., Zhang, P., Dai, X., Xiao, B., Yuan, L., Gao, J.: Focal
  self-attention for local-global interactions in vision transformers. NeurIPS
  (2021)

\bibitem{yang2021stacked}
Yang, L., Huang, Y., Sugano, Y., Sato, Y.: Stacked temporal attention:
  Improving first-person action recognition by emphasizing discriminative
  clips. BMVC  (2021)

\bibitem{ye2020few}
Ye, H.J., Hu, H., Zhan, D.C., Sha, F.: Few-shot learning via embedding
  adaptation with set-to-set functions. In: CVPR. pp. 8808--8817 (2020)

\bibitem{zeng2019graph}
Zeng, R., Huang, W., Tan, M., Rong, Y., Zhao, P., Huang, J., Gan, C.: Graph
  convolutional networks for temporal action localization. In: ICCV (2019)

\bibitem{zhang2020deepemd}
Zhang, C., Cai, Y., Lin, G., Shen, C.: Deepemd: Few-shot image classification
  with differentiable earth mover's distance and structured classifiers. In:
  CVPR (2020)

\bibitem{zhang2021temporal}
Zhang, C., Gupta, A., Zisserman, A.: Temporal query networks for fine-grained
  video understanding. In: CVPR (2021)

\bibitem{zhang2020few}
Zhang, H., Zhang, L., Qi, X., Li, H., Torr, P.H., Koniusz, P.: Few-shot action
  recognition with permutation-invariant attention. In: ECCV (2020)

\bibitem{zhang2021learning}
Zhang, S., Zhou, J., He, X.: Learning implicit temporal alignment for few-shot
  video classification. IJCAI  (2021)

\bibitem{zhou2018temporal}
Zhou, B., Andonian, A., Oliva, A., Torralba, A.: Temporal relational reasoning
  in videos. In: ECCV (2018)

\bibitem{zhu2018compound}
Zhu, L., Yang, Y.: Compound memory networks for few-shot video classification.
  In: ECCV (2018)

\bibitem{zhu2020label}
Zhu, L., Yang, Y.: Label independent memory for semi-supervised few-shot video
  classification. TPAMI  (2020)

\bibitem{zhu2021few}
Zhu, X., Toisoul, A., Perez-Rua, J.M., Zhang, L., Martinez, B., Xiang, T.:
  Few-shot action recognition with prototype-centered attentive learning. BMVC
  (2021)

\bibitem{zhu2021closer}
Zhu, Z., Wang, L., Guo, S., Wu, G.: A closer look at few-shot video
  classification: A new baseline and benchmark. BMVC  (2021)

\end{thebibliography}
\end{document}